\documentclass[runningheads]{llncs}
\usepackage{graphicx}

\usepackage{tikz}
\usepackage{comment}
\usepackage{amsmath,amssymb} 
\usepackage{color}

\usepackage[accsupp]{axessibility}  

\usepackage{multirow}
\usepackage{arydshln}
\usepackage{mathtools}
\usepackage{hyperref}
\usepackage{adjustbox}
\usepackage[utf8]{inputenc}
\begin{document}
\pagestyle{headings}
\mainmatter
\def\ECCVSubNumber{4}  

\title{TransNet: Category-Level \\Transparent Object Pose Estimation} 

\setlength\intextsep{\glueexpr\intextsep/2\relax}

\titlerunning{TransNet: Category-Level \\Transparent Object Pose Estimation}
%
\author{
Huijie Zhang
\and
Anthony Opipari
\and
Xiaotong Chen
\and
Jiyue Zhu
\and
Zeren Yu
\and
Odest Chadwicke Jenkins
}
\authorrunning{H. Zhang et al.}
%
\institute{University of Michigan, Ann Arbor MI 48109, USA\\
\email{\{huijiezh,topipari,cxt,cormaczh,yuzeren,ocj\}@umich.edu}
}
\maketitle

\begin{abstract}
Transparent objects present multiple distinct challenges to visual perception systems.
First, their lack of distinguishing visual features makes transparent objects harder to detect and localize than opaque objects. Even humans find certain transparent surfaces with little specular reflection or refraction, e.g. glass doors, difficult to perceive. A second challenge is that common depth sensors typically used for opaque object perception cannot obtain accurate depth measurements on transparent objects due to their unique reflective properties. Stemming from these challenges, we observe that transparent object instances within the same category (e.g. cups) look more similar to each other than to ordinary opaque objects of that same category. Given this observation, the present paper sets out to explore the possibility of category-level transparent object pose estimation rather than instance-level pose estimation. We propose \textit{\textbf{TransNet}}, a two-stage pipeline that learns to estimate category-level transparent object pose using localized depth completion and surface normal estimation. TransNet is evaluated in terms of pose estimation accuracy on a recent, large-scale transparent object dataset and compared to a state-of-the-art category-level pose estimation approach. Results from this comparison demonstrate that TransNet achieves improved pose estimation accuracy on transparent objects and key findings from the included ablation studies suggest future directions for performance improvements. The project webpage is available at: \href{https://progress.eecs.umich.edu/projects/transnet/}{https://progress.eecs.umich.edu/projects/transnet/}.

\end{abstract}

\begin{figure}[h!]
    \centering
    \includegraphics[width=\textwidth]{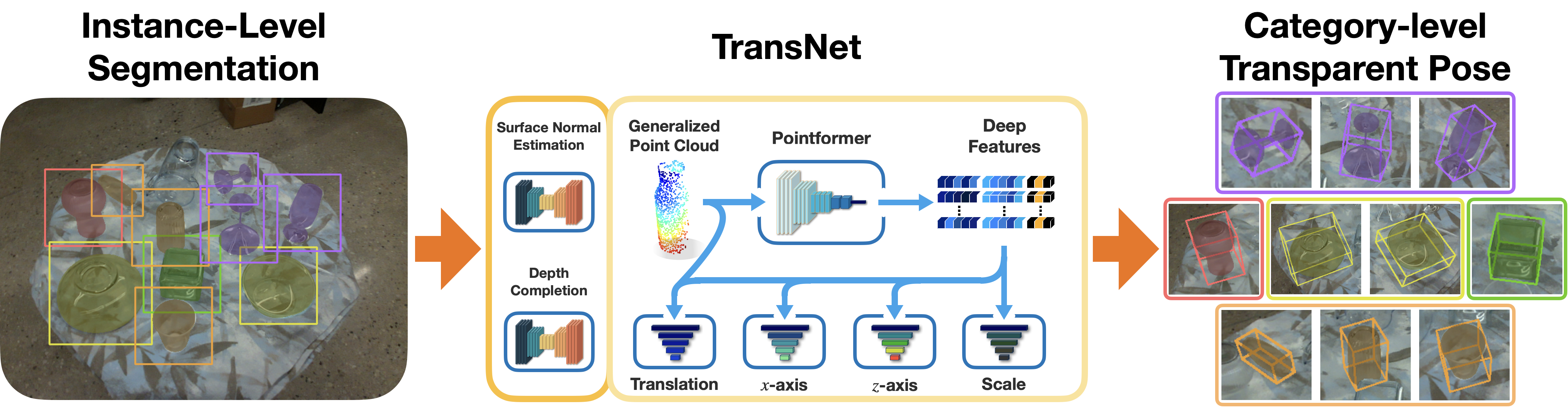}
    \caption{Overview of TransNet, a pipeline for category-level transparent object pose estimation. Given instance-level segmentation masks as input, TransNet estimates the 6 degrees of freedom pose and scale for each transparent object in the image. Internally, TransNet uses surface normal estimation, depth completion, and a transformer-based architecture for accurate pose estimation despite noisy sensor data.}
    \label{fig:teaser}
\end{figure}    

\section{Introduction}
\label{sec:intro}

\begin{figure}[h]
    \centering
    \includegraphics[width=0.8\textwidth]{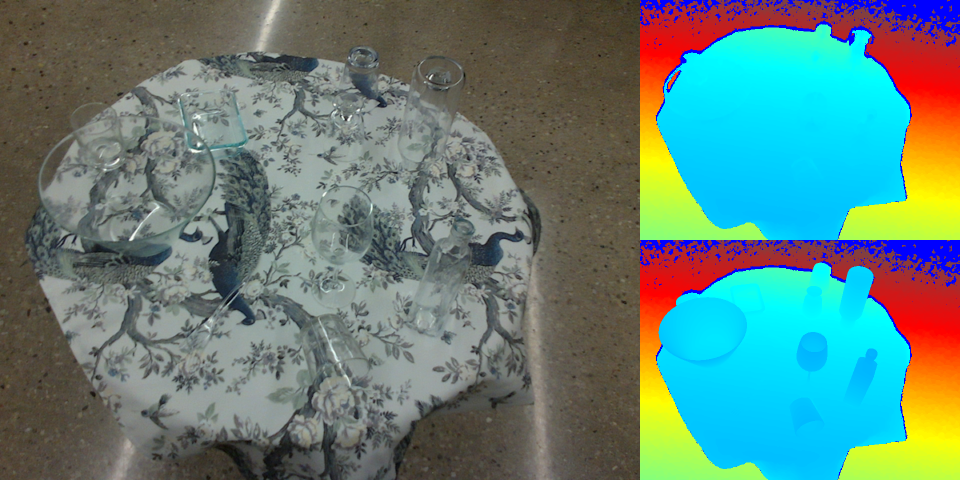}
    \caption{Challenge for transparent object perception. Images are from Clearpose dataset \cite{chen2022clearpose}. The left is an RGB image. The top right is the raw depth image and the bottom right is the ground truth depth image.}
    \label{fig:challenge}
\end{figure}

From glass doors and windows to kitchenware and all kinds of containers, transparent materials are prevalent throughout daily life.
Thus, perceiving the pose (position and orientation) of transparent objects is a crucial capability for autonomous perception systems seeking to interact with their environment.
However, transparent objects present unique perception challenges both in the RGB and depth domains.
As shown in Figure \ref{fig:challenge}, for RGB, the color appearance of transparent objects is highly dependent on the background, viewing angle, material, lighting condition, etc. due to light reflection and refraction effects.
For depth, common commercially available depth sensors record mostly invalid or inaccurate depth values within the region of transparency.
Such visual challenges, especially missing detection in the depth domain, pose severe problems for autonomous object manipulation and obstacle avoidance tasks.
This paper sets out to address these problems by studying how category-level transparent object pose estimation may be achieved using end-to-end learning.


Recent works have shown promising results on grasping transparent objects by completing the missing depth values followed by the use of a geometry-based grasp engine~\cite{sajjan2020clear,ichnowski2021dex,fang2022transcg}, or transfer learning from RGB-based grasping neural networks~\cite{weng2020multi}.
For more advanced manipulation tasks such as rigid body pick-and-place or liquid pouring, geometry-based estimations, such as symmetrical axes, edges~\cite{phillips2016seeing} or object poses~\cite{lysenkov2013recognition}, are required to model the manipulation trajectories.
Instance-level transparent object poses could be estimated from keypoints on stereo RGB images~\cite{liu2020keypose,liu2021stereobj} or directly from a single RGB-D image~\cite{xu20206dof} with support plane assumptions.
Recently emerged large-scale transparent object datasets~\cite{sajjan2020clear,xu2021seeing,liu2021stereobj,fang2022transcg,chen2022clearpose} pave the way for addressing the problem using deep learning.



In this work, we aim to extend the frontier of 3D transparent object perception with three primary contributions.
\begin{itemize}
    \item First, we explore the importance of depth completion and surface normal estimation in transparent object pose estimation. Results from these studies indicate the relative importance of each modality and their analysis suggests promising directions for follow-on studies.
    \item Second, we introduce \textit{TransNet}, a category-level pose estimation pipeline for transparent objects as illustrated in Figure \ref{fig:teaser}. It utilizes surface normal estimation, depth completion, and a transformer-based architecture to estimate transparent objects' 6D poses and scales.
    \item Third, we demonstrate that TransNet outperforms a baseline that uses a state-of-the-art opaque object pose estimation approach~\cite{di2022gpv} along with transparent object depth completion~\cite{fang2022transcg}.
\end{itemize}


\section{Related Works}
\label{sec:related_works}

\subsection{Transparent Object Visual Perception for Manipulation}

Transparent objects need to be perceived before being manipulated.
Lai \textit{et al.}~\cite{lai2015transparent} and Khaing \textit{et al.}~\cite{khaing2018transparent} developed CNN models to detect transparent objects from RGB images.
Xie \textit{et al.}~\cite{xie2020segmenting} proposed a deep segmentation model that achieved state-of-the-art segmentation accuracy.
ClearGrasp~\cite{sajjan2020clear} employed depth completion for use with pose estimation on robotic grasping tasks, where they trained three DeepLabv3+~\cite{chen2018encoder} models to perform image segmentation, surface normal estimation, and boundary segmentation.
Follow-on studies developed different approaches for depth completion, including implicit functions~\cite{zhu2021rgb}, NeRF features~\cite{ichnowski2021dex}, combined point cloud and depth features~\cite{xu2021seeing}, adversarial learning~\cite{tang2021depthgrasp}, multi-view geometry~\cite{chang2021ghostpose}, and RGB image completion~\cite{fang2022transcg}.
Without completing depth, Weng \textit{et al.}~\cite{weng2020multi} proposed a method to transfer the learned grasping policy from the RGB domain to the raw sensor depth domain.
For instance-level pose estimation, Xu \textit{et al.}~\cite{xu20206dof} utilized segmentation, surface normal, and image coordinate UV-map as input to a network similar to~\cite{tian2020robust} that can estimate 6 DOF object pose.
Keypose~\cite{liu2020keypose} was proposed to estimate 2D keypoints and regress object poses from stereo images using triangulation.
For other special sensors, Xu \textit{et al.}~\cite{xu2015transcut} used light-field images to do segmentation using a graph-cut-based approach.
Kalra \textit{et al.}~\cite{kalra2020deep} trained Mask R-CNN~\cite{he2017mask} using polarization images as input to outperform the baseline that was trained on only RGB images by a large margin.
Zhou \textit{et al.}~\cite{zhou2018plenoptic,zhou2019glassloc,zhou2020lit} employed light-field images to learn features for robotic grasping and object pose estimation.
Along with the proposed methods, massive datasets, across different sensors and both synthetic and real-world domains, have been collected and made public for various related tasks~\cite{xie2020segmenting,sajjan2020clear,liu2020keypose,zhou2020lit,kalra2020deep,liu2021stereobj,zhu2021rgb,xu2021seeing,fang2022transcg,chen2022clearpose}.
Compared with these previous works, and to the best of our knowledge we propose the first category-level pose estimation approach for transparent objects.
Notably, the proposed approach provides reliable 6D pose and scale estimates across instances with similar shapes.

\subsection{Opaque Object Category-level Pose Estimation}

Category-level object pose estimation is aimed at estimating unseen objects' 6D pose within seen categories, together with their scales or canonical shape.
To the best of our knowledge, there is not currently any category-level pose estimation works focusing on transparent objects, and the works mentioned below mostly consider opaque objects. They won't work well for transparency due to their dependence on accurate depth.
Wang \textit{et al.}~\cite{wang2019normalized} introduced the Normalized Object Coordinate Space (NOCS) for dense 3D correspondence learning, and used the Umeyama algorithm~\cite{umeyama1991least} to solve the object pose and scale.
They also contributed both a synthetic and a real dataset used extensively by the following works for benchmarking.
Later, Li \textit{et al.}~\cite{li2020category} extended the idea towards articulated objects.
To simultaneously reconstruct the canonical point cloud and estimate the pose, Chen \textit{et al.}~\cite{chen2020learning} proposed a method based on canonical shape space (CASS).
Tian \textit{et al.}~\cite{tian2020shape} learned category-specific shape priors from an autoencoder, and demonstrated its power for pose estimation and shape completion.
6D-ViT~\cite{zou20216d} and ACR-Pose~\cite{fan2021acr} extended this idea by utilizing pyramid visual transformer (PVT) and generative adversarial network (GAN)~\cite{goodfellow2014generative} respectively. 
Structure-guided prior adaptation (SGPA)~\cite{chen2021sgpa} utilized a transformer architecture for a dynamic shape prior adaptation.
Other than learning a dense correspondence, FS-Net~\cite{chen2021fs} regressed the pose parameters directly, and it proposed to learn two orthogonal axes for 3D orientation.
Also, it contributed to an efficient data augmentation process for depth-only approaches.
GPV-Pose~\cite{di2022gpv} further improved FS-Net by adding a geometric consistency loss between 3D bounding boxes, reconstruction, and pose. 
Also with depth as the only input, category-level point pair feature (CPPF)~\cite{you2022cppf} could reduce the sim-to-real gap by learning deep point pairs features.
DualPoseNet~\cite{lin2021dualposenet} benefited from rotation-invariant embedding for category-level pose estimation.
Differing from other works using segmentation networks to crop image patches as the first stage, CenterSnap~\cite{irshad2022centersnap} presented a single-stage approach for the prediction of 3D shape, 6D pose, and size.

Compared with opaque objects, we find the main challenge to perceive transparent objects is the poor quality of input depth.
Thus, the proposed TransNet takes inspiration from the above category-level pose estimation works regarding feature embedding and architecture design.
More specifically, TransNet leverages both Pointformer from PVT and the pose decoder from FS-Net and GPV-Pose.
In the following section, the TransNet architecture is described, focusing on how to integrate the single-view depth completion module and utilize imperfect depth predictions to learn pose estimates of transparent objects.
\section{TransNet}
\label{sec:method}

\begin{figure}[ht]
    \centering
    \includegraphics[width=\textwidth]{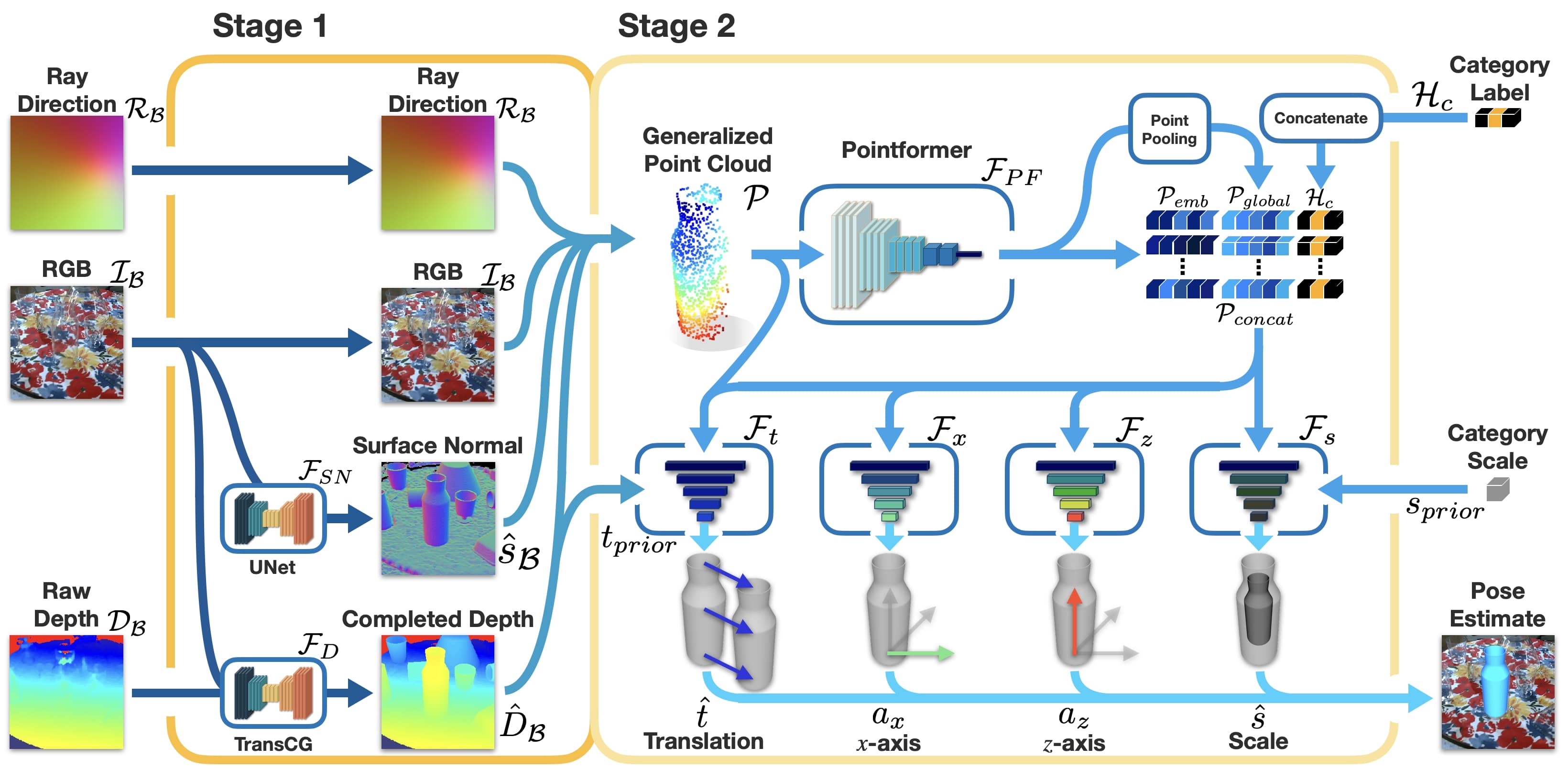}
    \caption{Architecture for TransNet. TransNet is a two-stage deep neural network for category-level transparent object pose estimation. The first stage uses an object instance segmentation (from Mask R-CNN~\cite{he2017mask}, which is not included in the diagram) to generate patches of RGB-D then used as input to a depth completion and a surface normal estimation network (RGB only). The second stage uses randomly sampled pixels within the objects' segmentation mask to generate a generalized point cloud formed as the per-pixel concatenation of ray direction, RGB, surface normal, and completed depth features. Pointformer~\cite{zou20216d}, a transformer-based point cloud embedding architecture, transforms the generalized point cloud into high-dimensional features. A concatenation of embedding features, global features, and a one-hot category label (from Mask R-CNN) is provided for the pose estimation module. The pose estimation module is composed of four decoders, one each for translation, $x$-axis, $z$-axis, and scale regression respectively. Finally, the estimated object pose is recovered and returned as output.}
    \label{fig:arch}
\end{figure}    

Given an input RGB-D pair ($\mathcal{I}$, $\mathcal{D}$), our goal is to predict objects' 6D rigid body transformations $[\textbf{R}|\textbf{t}]$ and 3D scales $\textbf{s}$ in the camera coordinate frame, where $\textbf{R} \in SO(3), \textbf{t} \in \mathbb{R}^{3}$ and $\textbf{s} \in \mathbb{R}^{3}_{+}$. 
In this problem, inaccurate/invalid depth readings exist within the image region corresponding to transparent objects (represented as a binary mask $\mathcal{M}_t$).
To approach the category-level pose estimation problem along with inaccurate depth input, we propose a novel two-stage deep neural network pipeline, called \textbf{TransNet}.

\subsection{Architecture Overview}

Following recent work in object pose estimation~\cite{wang2019densefusion,chen2021fs,di2022gpv}, we first apply a pre-trained instance segmentation module (Mask R-CNN \cite{he2017mask}) that has been fine-tuned on the pose estimation dataset to extract the objects' bounding box patches, masks, and category labels to separate the objects of interest from the entire image.

The first stage of TransNet takes the patches as input and attempts to correct the inaccurate depth posed by transparent objects. Depth completion (TransCG \cite{fang2022transcg}) and surface normal estimation (U-Net \cite{ronneberger2015u}) are applied on RGB-D patches to obtain estimated depth-normal pairs.
The estimated depth-normal pairs, together with RGB and ray direction patches, are concatenated to feature patches, followed by a random sampling strategy within the instance masks to generate generalized point cloud features.

In the second stage of TransNet, the generalized point cloud is processed through Pointformer \cite{zou20216d}, a transformer-based point cloud embedding module, to produce concatenated feature vectors. The pose is then separately estimated in four decoder modules for object translation, $x$-axis, $z$-axis, and scale respectively.
The estimated rotation matrix can be recovered using the estimated two axes. Each component is discussed in more detail in the following sections.

\subsection{Object Instance Segmentation}

Similar to other categorical pose estimation work \cite{di2022gpv}, we train a Mask R-CNN \cite{he2017mask} model on the same dataset used for pose estimation to obtain the object's bounding box $\mathcal{B}$, mask $\mathcal{M}$ and category label $\mathcal{H}_c$.
Patches of ray direction $\mathcal{R}_{\mathcal{B}}$, RGB $\mathcal{I}_{\mathcal{B}}$ and raw depth $\mathcal{D}_{\mathcal{B}}$ are extracted from the original data source following bounding box $\mathcal{B}$, before inputting to the first stage of TransNet.

\subsection{Transparent object depth completion}

Due to light reflection and refraction on transparent material, the depth of transparent objects is very noisy.
Therefore, depth completion is necessary to reduce the sensor noise.
Given the raw RGB-D patch ($\mathcal{I}_{\mathcal{B}}$, $\mathcal{D}_{\mathcal{B}}$) pair and transparent mask $\mathcal{M}_t$ (a intersection of transparent objects' masks within bounding box $\mathcal{B}$), transparent object depth completion $\mathcal{F}_{D}$ is applied to obtain the completed depth of the transparent region $\{\hat{\mathcal{D}}_{(i, j)}|(i, j)\in\mathcal{M}_t \}$.

Inspired by one state-of-the-art depth completion method, TransCG \cite{fang2022transcg}, we incorporate a similar multi-scale depth completion architecture into TransNet.
\begin{equation}
\label{eq:dc}
    \hat{\mathcal{D}}_\mathcal{B} = \mathcal{F}_{D}\left(\mathcal{I}_\mathcal{B}, \mathcal{D}_\mathcal{B}\right)
\end{equation}

We use the same training loss as TransCG:
\begin{align}
\begin{split}
    &\mathcal{L} = \mathcal{L}_d + \lambda_{smooth} \mathcal{L}_s \\
    &\mathcal{L}_d = \frac{1}{N_p}\sum_{p\in\mathcal{M}_t \bigcap \mathcal{B}}\left\lVert\hat{\mathcal{D}}_p -  \mathcal{D}^{*}_p\right\rVert^2 \\
    &\mathcal{L}_s = \frac{1}{N_p}\sum_{p\in\mathcal{M}_t \bigcap \mathcal{B}}\left(1 - \text{cos}\left\langle\mathcal{N}(\hat{\mathcal{D}}_p), \mathcal{N}(\mathcal{D}^{*}_p)\right\rangle\right)
\end{split}
\end{align}

where $\mathcal{D}^{*}$ is the ground truth depth image patch, $p\in\mathcal{M}_t \bigcap \mathcal{B}$ represents the transparent region in the patch, $\left\langle\boldsymbol{\cdot \; , \; \cdot}\right\rangle$ denotes the dot product operator and $\mathcal{N}(\boldsymbol{\cdot})$ denotes the operator to calculate surface normal from depth. $\mathcal{L}_d$ is $L_2$ distance between estimated and ground truth depth within the transparency mask. $\mathcal{L}_s$ is the cosine similarity between surface normal calculated from estimated and ground truth depth. $\lambda_{smooth}$ is the weight between the two losses. 

\subsection{Transparent object surface normal estimation}

Surface normal estimation $\mathcal{F}_{SN}$ estimates surface normal $\mathcal{S}_{\mathcal{B}}$ from RGB image $\mathcal{I}_{\mathcal{B}}$. Although previous category-level pose estimation works \cite{di2022gpv,chen2021fs} show that depth is enough to obtain opaque objects' pose, experiments in Section \ref{exp:generalized point cloud} demonstrate that surface normal is not a redundant input for transparent object pose estimation. Here, we slightly modify U-Net \cite{ronneberger2015u} to perform the surface normal estimation. 
\begin{equation}
\label{eq:sne}
    \hat{\mathcal{S}}_\mathcal{B} = \mathcal{F}_{SN}\left(\mathcal{I}_\mathcal{B}\right)
\end{equation}

We use the cosine similarity loss:
\begin{align}
\begin{split}
    &\mathcal{L} = \frac{1}{N_p}\sum_{p\in\mathcal{B}}\left(1 - \text{cos}\left\langle\hat{\mathcal{S}}_p, \mathcal{S}^{*}_p\right\rangle\right)
\end{split}
\end{align}

where $p\in \mathcal{B}$ means the loss is applied for all pixels in the bounding box $\mathcal{B}$. 

\subsection{Generalized point cloud}

As input to the second stage, generalized point cloud $\mathcal{P}\in \mathbb{R}^{N\times d}$ is a stack of $d$-dimensional features from the first stage taken at $N$ sample points, inspired from \cite{xu20206dof}. To be more specific, $d=10$ in our work. Given the completed depth $\hat{\mathcal{D}}_\mathcal{B}$ and predicted surface normal $\hat{\mathcal{S}}_\mathcal{B}$ from Equation \eqref{eq:dc}, \eqref{eq:sne}, together with RGB patch $\mathcal{I}_\mathcal{B}$ and ray direction patch $\mathcal{R}_\mathcal{B}$, a concatenated feature patch is given as $\left[\mathcal{I}_\mathcal{B},  \hat{\mathcal{D}}_\mathcal{B},  \hat{\mathcal{S}}_\mathcal{B},  \mathcal{R}_\mathcal{B}\right] \in \mathbb{R}^{H \times W \times 10}$.
Here the ray direction $\mathcal{R}$ represents the direction from camera origin to each pixel in the camera frame. For each pixel $(u,v)$:
\begin{align}
\begin{split}
    &p = \begin{bmatrix}u & v & 1\end{bmatrix}^T \\
    &\mathcal{R} = \frac{K^{-1} p}{\left\lVert K^{-1} p\right\rVert^2}
\end{split}
\end{align}

where $p$ is the homogeneous UV coordinate in the image plane and $K$ is the camera intrinsic.
The UV mapping itself is an important cue when estimating poses from patches \cite{jiang2022uni6d}, as it provides information about the relative position and size of the patches within the overall image.
We use ray direction instead of UV mapping because it also contains camera intrinsic information.

We randomly sample $N$ pixels within the transparent mask of the feature patch to obtain the generalized point cloud $\mathcal{P}\in \mathbb{R}^{N\times 10}$. A more detailed experiment in Section \ref{exp:generalized point cloud} explores the best choice of the generalized point cloud.

\subsection{Transformer Feature embedding}

Given generalized point cloud $\mathcal{P}$, we apply an encoder and multi-head decoder strategy to get objects' poses and scales. We use Pointformer~\cite{zou20216d}, a multi-stage transformer-based point cloud embedding method:
\begin{equation}
\label{eq:emb}
    \mathcal{P}_{emb} = \mathcal{F}_{PF}\left(\mathcal{P}\right)
\end{equation}

where $\mathcal{P}_{emb} \in \mathbb{R}^{N\times d_{emb}}$ is a high-dimensional feature embedding.
During our experiments, we considered other common point cloud embedding methods such as 3D-GCN~\cite{Lin_2020_3dgcn} demonstrating their power in many category-level pose estimation methods \cite{chen2021fs,di2022gpv}.
During feature aggregation for each point, they use the nearest neighbor algorithm to search nearby points within coordinate space, then calculate new features as a weighted sum of the features within surrounding points. Due to the noisy input $\hat{D}$ from Equation \eqref{eq:dc}, the nearest neighbor may become unreliable by producing noisy feature embeddings. On the other hand, Pointformer aggregates feature by a transformer-based method. The gradient back-propagates through the whole point cloud. More comparisons and discussions in Section \ref{exp:embedding} demonstrate that transformer-based embedding methods are more stable than nearest neighbor-based methods when both are trained on noisy depth data.

Then we use a Point Pooling layer (a multilayer perceptron (MLP) plus max-pooling) to extract the global feature $\mathcal{P}_{global}$, and concatenate it with local feature $\mathcal{P}_{emb}$ and the one-hot category $\mathcal{H}_{c}$ label from instance segmentation for the decoder: 
\begin{align}
\begin{split}
\label{eq:concat}
    &\mathcal{P}_{global} = \text{MaxPool}\left(\text{MLP}\left(\mathcal{P}_{emb}\right)\right) \\
    &\mathcal{P}_{concat} = \left[\mathcal{P}_{emb}, \mathcal{P}_{global}, \mathcal{H}_{c}\right]
\end{split}
\end{align}

\subsection{Pose and Scale Estimation}

After we extract the feature embeddings from multi-modal input, we apply four separate decoders for translation, $x$-axis, $z$-axis, and scale estimation.

\noindent\textbf{Translation Residual Estimation} As demonstrated in \cite{chen2021fs}, residual estimation achieves better performance than direct regression by learning the distribution of the residual between the prior and actual value. The translation decoder $\mathcal{F}_{t}$ learns a 3D translation residual from the object translation prior $t_{prior}$ calculated as the average of predicted 3D coordinate over the sampled pixels in $\mathcal{P}$. To be more specific:
\begin{align}
\begin{split}
\label{eq:trans}
    &t_{prior} = \frac{1}{N_p}\sum_{p\in N} K ^{-1} \left[u_p \  v_p \  1\right]^T \hat{\mathcal{D}_p} \\
    &\hat{t} = t_{prior} + \mathcal{F}_{t}\left(\left[\mathcal{P}_{concat}, \mathcal{P}\right]\right) \\
\end{split}
\end{align}
Where $K$ is the camera intrinsic and $u_p$, $v_p$ are the 2D pixel coordinate for the selected pixel. We also use the $L_1$ loss between the ground truth and estimated position:
\begin{equation}
\label{eq:trans_loss}
    \mathcal{L}_t = \left\lvert\hat{t} - t^*\right\rvert
\end{equation}

\noindent\textbf{Pose Estimation} Similar to \cite{chen2021fs}, rather than directly regress the rotation matrix $R$, it is more effective to decouple it into two orthogonal axes and estimate them separately.
As shown in Figure \ref{fig:arch}, we decouple $R$ into the $z$-axis $a_z$ (red axis) and $x$-axis $a_x$ (green axis).
Following the strategy of confidence learning in \cite{di2022gpv}, the network learns confidence values to deal with the problem that the regressed two axes are not orthogonal:
\begin{align}
\begin{split}
\label{eq:rot}
    &\left[\hat{a}_i, c_i\right] = \mathcal{F}_i\left(\mathcal{P}_{concat}\right), \  i\in \left\{x, z\right\} \\
    &\theta_z = \frac{c_x}{c_x + c_z}\left(\theta - \frac{\pi}{2}\right)\\
    &\theta_x = \frac{c_z}{c_x + c_z}\left(\theta - \frac{\pi}{2}\right)
\end{split}
\end{align}
where $c_x, c_z$ denote the confidence for the learned axes. $\theta$ represents the angle between $a_x$ and $a_z$.
$\theta_x, \theta_z$ are obtained by solving an optimization problem and then used to rotate the $a_x$ and $a_z$ within their common plane.
More details can be found in \cite{di2022gpv}.
For the training loss, first, we use $L_1$ loss and cosine similarity loss for axis estimation:
\begin{align}
\begin{split}
\label{eq:rot_loss}
    &\mathcal{L}_{r_i} = \left\lvert\hat{a}_i - a^*_i\right\rvert + 1 - \left\langle\hat{a}_i, a^*_i\right\rangle, \  i\in \left\{x, z\right\} 
\end{split}
\end{align}

Then to constrain the perpendicular relationship between two axes, we add the angular loss:
\begin{align}
\begin{split}
\label{eq:rot_ang_loss}
    &\mathcal{L}_a = \left\langle\hat{a}_x, \hat{a}_z\right\rangle
\end{split}
\end{align}

To learn the axis confidence, we add the confidence loss, which is the $L_1$ distance between estimated confidence and exponential $L_2$ distance between the ground truth and estimated axis:
\begin{align}
\begin{split}
\label{eq:rot_con_loss}
    &\mathcal{L}_{con_i} = \left\lvert c_i - \text{exp}\left(\alpha \left\lVert \hat{a}_i - a^*_i \right\rVert_2\right)\right\rvert, \  i\in \left\{x, z\right\} 
\end{split}
\end{align}

where $\alpha$ is a constant to scale the distance. 

Thus the overall loss for the second stage is:
\begin{align}
\begin{split}
\label{eq:all_loss}
    \mathcal{L} = &\lambda_s\mathcal{L}_s + \lambda_t\mathcal{L}_t + \lambda_{r_x}\mathcal{L}_{r_x} + \lambda_{r_z}\mathcal{L}_{r_z} + \\
    &\lambda_{r_a}\mathcal{L}_{a} + \lambda_{con_x}\mathcal{L}_{con_x} + \lambda_{con_z}\mathcal{L}_{con_z}
\end{split}
\end{align}

To deal with object symmetry, we apply specific treatments for different symmetry types.
For axial symmetric objects (those that remain the same shape when rotating around one axis), we ignore the loss for the $x$-axis, $i.e., \mathcal{L}_{con_x}, \mathcal{L}_{r_x}$.
For planar symmetric objects (those that remain the same shape when mirrored about one or more planes), we generate all candidate $x$-axis rotations. 
For example, for an object symmetric about the $x-z$ plane and $y-z$ plane, rotating the $x$-axis about the $z$-axis by $\pi$ radians will not affect the object's shape.
The new $x$-axis is denoted as $a_{x_{\pi}}$ and the loss for the $x$-axis is defined as the minimum loss of both candidates: 
\begin{align}
\begin{split}
\label{eq:rot_plannar}
    \mathcal{L}_x = \text{min}\left(\mathcal{L}_x(a_x), \mathcal{L}_x(a_{x_{\pi}})\right)
\end{split}
\end{align}

\noindent\textbf{Scale Residual Estimation}  
Similar to the translation decoder, we define the scale prior $s_{prior}$ as the average of scales of all object 3D CAD models within each category. Then the scale of a given instance is calculated as follows:
\begin{align}
\begin{split}
\label{eq:scale}
    &\hat{s} = s_{prior} + \mathcal{F}_{s}\left(\mathcal{P}_{concat}\right) \\
\end{split}
\end{align}

 The loss function is defined as the $L_1$ loss between the ground truth scale and estimated scale:

\begin{equation}
\label{eq:scale_loss}
    \mathcal{L}_s = \left\lvert\hat{s} - s^*\right\rvert
\end{equation}
\section{Experiments}
\label{sec:result}


\textbf{Dataset} We evaluated TransNet and baseline models on the Clearpose Dataset \cite{chen2022clearpose} for categorical transparent object pose estimation.
The Clearpose Dataset contains over 350K real-world labeled RGB-D frames in 51 scenes, 9 sets, and around 5M instance annotations covering 63 household objects.
We selected 47 objects and categorize them into 6 categories, \textit{bottle, bowl, container, tableware, water cup, wine cup.}
We used all the scenes in set2, set4, set5, and set6 for training and scenes in set3 and set7 for validation and testing.
The division guaranteed that there were some unseen objects for testing within each category.
Overall, we used 190K images for training and 6K for testing.
For training depth completion and surface normal estimation, we used the same dataset split.

\noindent\textbf{Implementation Details} Our model was trained in several stages. For all the experiments in this paper, we were using the ground truth instance segmentation as input, which could also be obtained by Mask R-CNN \cite{he2017mask}. 
The image patches were generated from object bounding boxes and re-scaled to a fixed shape of $256\times256$ pixels. For TransCG, we used AdamW optimizer~\cite{loshchilov2017decoupled} for training with $\lambda_{smooth} = 0.001$ and the overall learning rate is $0.001$ to train the model till converge. 
For U-Net, we used the Adam optimizer~\cite{kingma2014adam} with a learning rate of $1e^{-4}$ to train the model until convergence.
For both surface normal estimation and depth completion, the batch size was set to 24 images.
The surface normal estimation and depth completion model were frozen during the training of the second stage.

For the second stage, the training hyperparameters for Pointformer followed those used in \cite{zou20216d}.
We used data augmentation for RGB features and instance mask for sampling generalized point cloud.
A batch size of 18 was used. To balance sampling distribution across categories, 3 instance samples were selected randomly for each of 6 categories.
We followed GPV-Pose \cite{di2022gpv} on training hyper-parameters.
The learning rate for all loss terms were kept the same during training, $\left\{\lambda_{r_x}, \lambda_{r_z}, \lambda_{r_a}, \lambda_{t}, \lambda_{s}, \lambda_{con_x}, \lambda_{con_z}\right\} = \left\{8, 8, 4, 8, 8, 1, 1\right\}\times0.0001$.
We used the Ranger optimizer \cite{liu2019variance,yong2020gradient,zhang2019lookahead} and used a linear warm-up for the first 1000 iterations, then used a cosine annealing method at the 0.72 anneal point.
All the experiments for pose estimation were trained on a 16G RTX3080 GPU for 30 epochs with 6000 iterations each. All the categories were trained on the same model, instead of one model per category. 


\noindent\textbf{Evaluation metrics} For category-level pose estimation, we followed \cite{di2022gpv,chen2021fs} using 3D intersection over union (IoU) between the ground truth and estimated 3D bounding box (we used the estimated scale and pose to draw an estimated 3D bounding box) at 25\%, 50\% and 75\% thresholds. Additionally, we used $5^{\circ}2cm$, $5^{\circ}5cm$, $10^{\circ}5cm$, $10^{\circ}10cm$ as metrics. The numbers in the metrics represent the percentage of the estimations with errors under such degree and distance. For section \ref{exp:depth-surfacenormal}, we also used separated translation and rotation metrics: $2cm$, $5cm$, $10cm$, $5^{\circ}$, $10^{\circ}$ that calculate percentage with respect to one factor. 

For depth completion evaluation, we calculated the root of mean squared error (RMSE), absolute relative error (REL) and mean absolute error (MAE), and used $\delta_{1.05}$, $\delta_{1.10}$, $\delta_{1.25}$ as metrics, while $\delta_n$ was calculated as:

\begin{equation}
    \delta_n = \frac{1}{N_p}\sum_{p}\textbf{I}\left(\text{max}\left(\frac{\hat{\mathcal{D}}_p}{\mathcal{D}^*_p}, \frac{\mathcal{D}^*_p}{\hat{\mathcal{D}}_p}\right) < n\right)
\end{equation}

\noindent where $\textbf{I}(\boldsymbol{\cdot})$ represents the indicator function. $\hat{\mathcal{D}_p}$ and $\mathcal{D}^*_p$ mean estimated and ground truth depth for each pixel $p$.

For surface normal estimation, we calculated RMSE and MAE errors and used $11.25^{\circ}$, $22.5^{\circ}$, and $30^{\circ}$ as thresholds. Here $11.25^{\circ}$ represents the percentage of estimates with an angular distance less than $11.25^{\circ}$ from ground truth surface normal.

\subsection{Comparison with Baseline}
\label{exp:baseline}
\setlength{\tabcolsep}{1pt}
\begin{table}
\begin{center}
\caption{Comparison with the baseline on the Clearpose Dataset.}
\label{table:baseline}
\begin{tabular}{c|ccccccc}
\hline
Method & $\text{3D}_{25}$$\uparrow$ & $\text{3D}_{50}$$\uparrow$ & $\text{3D}_{75}$$\uparrow$ & $5^{\circ}2\text{cm}$$\uparrow$ & $5^{\circ}5\text{cm}$$\uparrow$ & $10^{\circ}5\text{cm}$$\uparrow$ & $10^{\circ}10\text{cm}$$\uparrow$\\
\hline
GPV-Pose &\textbf{93.7}  & 58.3 & 10.5 & 0.4 & 1.5 & 7.4 & 9.1 \\
\hline
TransNet &90.3 & \textbf{67.4} & \textbf{22.1} & \textbf{2.4} & \textbf{7.5} & \textbf{23.6} & \textbf{27.6} \\
\hline
\end{tabular}
\end{center}
\end{table}
\setlength{\tabcolsep}{1.4pt}

\begin{figure}[ht]
    \centering
    \includegraphics[width=\textwidth]{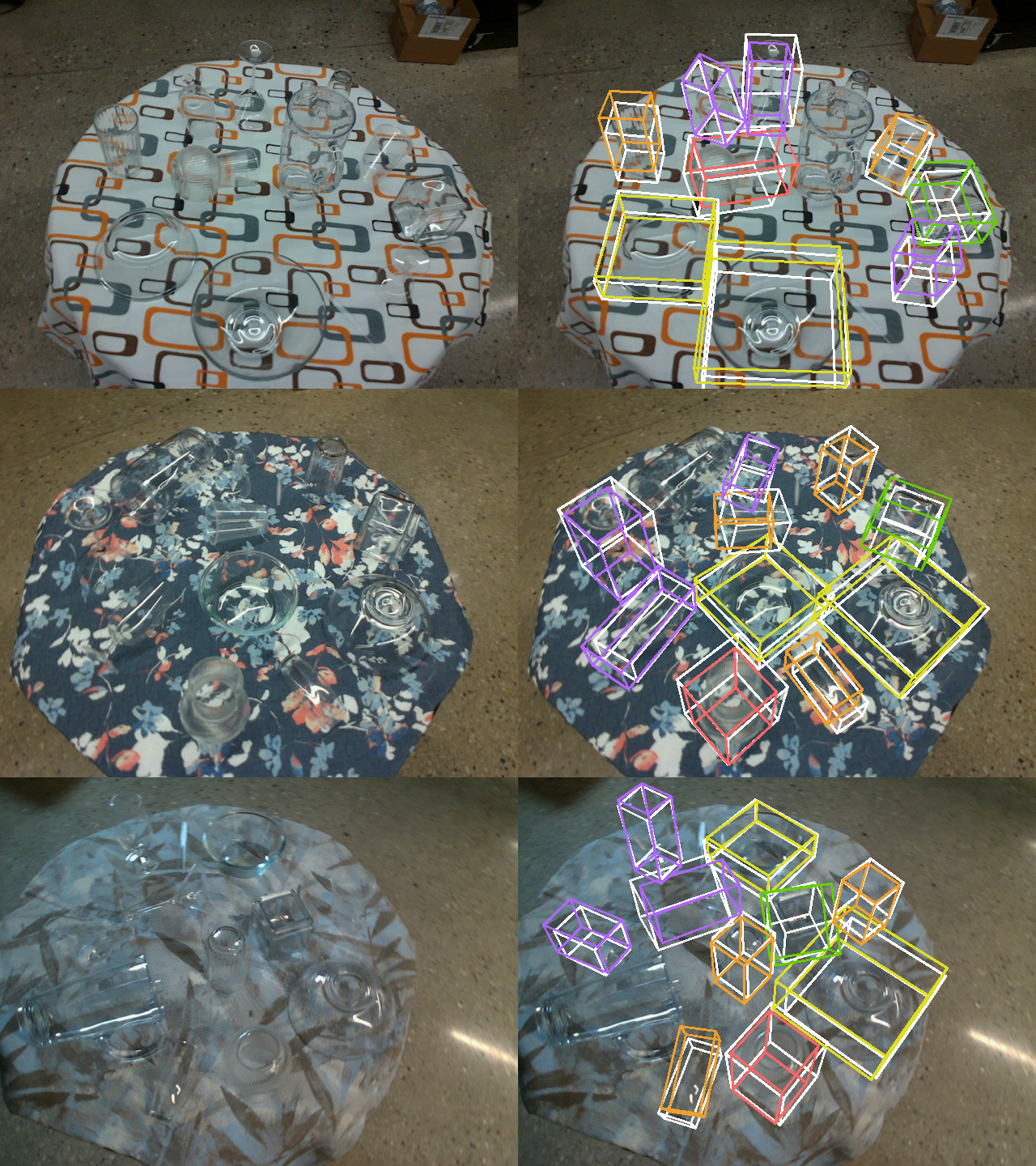}
    \caption{Qualitative results of category-level pose estimates from TransNet. The left column is the original RGB image within our test set and the right column is the pose estimation results. The white bounding box is the ground truth and the colored one is the estimation result. Different colors represent different categories. For axial symmetric objects, because we only care about the scale and z-axis, we use the ground truth x-axis and estimated z-axis to calculate the estimated x-axis, for better visualization. In the figure, there is a pitcher without either ground truth or estimated bounding box because it is not within any of the defined categories, so we ignore it for both training and testing.}
    \label{fig:visual}
\end{figure}

We chose one state-of-the-art categorical opaque object pose estimation model (GPV-Pose \cite{di2022gpv}) as a baseline, which was trained with estimated depth from TransCG \cite{fang2022transcg} for a fair comparison. From Table \ref{table:baseline}, TransNet outperformed the baseline in most of the metrics on the Clearpose dataset. $3\text{D}_{25}$ is very easy to learn, so there is no huge difference between them. For the rest of the metrics, TransNet achieved around 2$\times$ the percentage on $3\text{D}_{50}$, 3$\times$ on $10^{\circ}5\text{cm}, 10^{\circ}10\text{cm}$ and 5$\times$ on $5^{\circ}5\text{cm}, 5^{\circ}2\text{cm}$ over the baseline. Qualitative results are shown in Figure \ref{fig:visual} for TransNet. 

\subsection{Embedding method analysis}
\label{exp:embedding}

In Table \ref{table:emb_ab}, we compared the embedding method between 3D-GCN \cite{Lin_2020_3dgcn} and Pointformer \cite{zou20216d} on TransNet. Modalities for generalized point cloud were depth, RGB and ray direction (without surface normal) for all the trials. The only differences between them were depth type and embedding methods. With ground truth input, 3D-GCN and Pointformer achieved similar results. For some metrics, \textit{i.e.} $5^{\circ}5cm$, 3D-GCN was even better.
But when the ground truth depth was changed to estimated depth (modeling the change from opaque to transparent setting), Pointformer retained much more accuracy than 3D-GCN. Here is our explanation. Like many point cloud embedding methods, 3D-GCN propagates information between nearest neighbors. It is a very efficient method given a point cloud with low noise. But given the completed depth, high noise makes it unstable to pass data among neighbors. While for Pointformer, information is passed through the whole point cloud, no matter how large the noise is. Therefore, given depth information with large uncertainty, the transformer-based embedding method might be more powerful than embedding methods using nearest neighbors. 


\setlength{\tabcolsep}{1pt}
\begin{table}
\begin{center}
\caption{Comparison between different embedding methods}
\label{table:emb_ab}
\begin{tabular}{c|c|ccccccc}
\hline
Depth type & Embedding& $\text{3D}_{25}$$\uparrow$ & $\text{3D}_{50}$$\uparrow$ & $\text{3D}_{75}$$\uparrow$ & $5^{\circ}2\text{cm}$$\uparrow$ & $5^{\circ}5\text{cm}$$\uparrow$ & $10^{\circ}5\text{cm}$$\uparrow$ & $10^{\circ}10\text{cm}$$\uparrow$\\
\hline
\multirow{2}{*}{Ground truth} 
& 3D-GCN & \textbf{90.0} & \textbf{84.1} & 43.0 & 21.4 & \textbf{48.0} & \textbf{61.8} & \textbf{64.7} \\
& Pointformer & \textbf{90.0} & 81.8 & \textbf{56.5} & \textbf{24.1} & 39.3 & 59.0 & 60.7 \\
\hline
\multirow{2}{*}{Estimation} 
& 3D-GCN & \textbf{88.8}  & 59.8 & 10.4 & 0.9 & 3.4 & 12.3 & 15.4 \\
& Pointformer &88.5 & \textbf{62.2} & \textbf{17.6} & \textbf{1.6} & \textbf{5.0} & \textbf{17.4} & \textbf{20.9} \\
\hline
\end{tabular}
\end{center}
\end{table}
\setlength{\tabcolsep}{1.4pt}

\subsection{Ablation study of generalized point cloud}
\label{exp:generalized point cloud}
We explored different combinations of feature inputs for the generalized point cloud to find the one most suitable for TransNet. Results are shown in Table \ref{table:gpcd_ab}. 
For trials 1 and 2, we compared the effect of adding estimated surface normal to the generalized point cloud. All the metrics demonstrated that the inclusion of surface normal does improve the resulting pose estimation accuracy. 
\begin{table}
\begin{center}
\caption{Ablation study for a different combination of the generalized point cloud. 
For both trials, we also use RGB as an input feature for the generalized point cloud.}
\label{table:gpcd_ab}
\begin{adjustbox}{width=\columnwidth,center}
\begin{tabular}{c|ccc|ccccccc}
\hline
Trial & depth & normal & ray-direction &
$3D_{25}$$\uparrow$ & $3D_{50}$$\uparrow$ & $3D_{75}$$\uparrow$ & $5^{\circ}2cm$$\uparrow$ & $5^{\circ}5cm$$\uparrow$ & $10^{\circ}5cm$$\uparrow$ & $10^{\circ}10cm$$\uparrow$\\
\hline
1 & \checkmark & & \checkmark 
& 88.5 & 62.2 & 17.6 & 1.6 & 5.0 & 17.4 & 20.9 \\
2 & \checkmark & \checkmark & \checkmark 
& \textbf{90.3} & \textbf{67.4} & \textbf{22.1} & \textbf{2.4} & \textbf{7.5} & \textbf{23.6} & \textbf{27.6} \\
\hline
\end{tabular}
\end{adjustbox}
\end{center}
\end{table}

\subsection{Depth and surface normal exploration on TransNet}
\label{exp:depth-surfacenormal}

\setlength{\tabcolsep}{1pt}
\begin{table}
\begin{center}
\caption{Accuracy for depth completion on Clearpose dataset. 
All the metrics are calculated within the transparent mask.}
\label{table:dc}
\begin{tabular}{c|cccccc}
\hline
Metric &RMSE$\downarrow$ & REL$\downarrow$ & MAE$\downarrow$ & $\delta_{1.05}$$\uparrow$ & $\delta_{1.10}$$\uparrow$ & $\delta_{1.25}$$\uparrow$\\
\hline
\text{Value} & 0.055 & 0.044 & 0.041 & 68.93 & 89.40 & 98.93 \\
\hline
\end{tabular}

\bigskip

\caption{Accuracy for surface normal estimation on Clearpose dataset.
}
\label{table:sn}
\begin{tabular}{c|cccccc}
\hline
Metric & RMSE$\downarrow$ & MAE$\downarrow$ & $11.25^{\circ}$$\uparrow$ & $22.5^{\circ}$$\uparrow$ & $30^{\circ}$$\uparrow$\\
\hline
\text{Value} & 0.1915 & 0.1334 & 56.75 & 88.45 & 96.64 \\
\hline
\end{tabular}

\bigskip

\caption{Evaluation for depth and surface normal accuracy on TransNet. 
}
\label{table:d&sn_ab}
\begin{adjustbox}{width=\columnwidth,center}
\begin{tabular}{c|cc|cccccccccccc}
\hline
Trial & Depth & Normal & $3D_{25}$$\uparrow$ & $3D_{50}$$\uparrow$ & $3D_{75}$$\uparrow$ & $5^{\circ}2cm$$\uparrow$ & $5^{\circ}5cm$$\uparrow$ & $10^{\circ}5cm\uparrow$ & $10^{\circ}10cm$$\uparrow$ & $5^{\circ}\uparrow$ & $10^{\circ}\uparrow$ & $2cm\uparrow$ & $5cm\uparrow$ & $10cm\uparrow$ \\
\hline
1 & GT  & GT & \textbf{95.1} & \textbf{87.7} & \textbf{66.7} & \textbf{31.8} & \textbf{48.4} & \textbf{66.5} & \textbf{66.7} & \textbf{47.3} & \textbf{66.3} & \textbf{63.3} & \textbf{97.9} & \textbf{99.9} \\
2 & GT & EST & 90.9 & 82.1 & 56.3 & 23.4 & 36.5 & 58.0 & 59.6 & 37.3 & 59.6 & 53.6 & 97.2 & 99.9\\
3 & EST & GT & 94.0 & 83.8 & 34.3 & 8.1 & 29.9 & 47.8 & 60.3 & 37.3 & 61.8 & 22.2 & 77.1 & 97.4\\
4 & EST & EST & 90.3 & 67.4 & 22.1 & 2.4 & 7.5 & 23.6 & 27.6 & 8.8 & 28.1 & 16.6 & 77.4 & 96.8\\
\hline
\end{tabular}
\end{adjustbox}
\end{center}
\end{table}
\setlength{\tabcolsep}{1.4pt}

We explored the combination of depth and surface normal with different accuracy. Results in Table \ref{table:dc} and Table \ref{table:sn} show performance for TransCG and U-Net separately. ``GT" and ``EST" in Table~\ref{table:d&sn_ab} represent ground truth and estimated input for depth and surface normal respectively. From the comparison of results among trials 1 - 3, accurate depth is more essential than surface normal for category-level transparent object pose estimation. For instance, as the ground truth depth changes to the estimated depth from trial 1 to trial 3, $5^{\circ}2cm$ decreases by 23.7. Compared with surface normal estimation, $5^{\circ}2cm$ only decreases by 8.4 between trial 1 and trial 2. 
More specifically, from decoupled rotation and translation metrics, we can see that $2cm$ decreases by 41.1 between trial 1 and trial 3 compared to 9.7 between trial 1 and trial 2, meaning that depth accuracy is more important for translation estimation. Focusing on $2cm, 5cm, 10cm$ between trial 1 and trial 4, the first metric decreases by 46.7 but the latter two lose much less (20.5 for $5cm$ and 3.1 for $10cm$). This can be explained by the result of depth completion accuracy shown in Table \ref{table:dc} (MAE = 0.041m, between $2cm$ and $5cm$). 
From the comparison of trial 1-4 on metrics $5^{\circ}$ and $10^{\circ}$, we can see that either accurate surface normal or accurate depth can support good performance in rotation metrics (for either trial 2 or trial 3, $5^{\circ}$ decreases by 10.0 and $10^{\circ}$ decreased by around 7). Once we use the estimation version of both, $5^{\circ}$ decreases by 38.5 and $10^{\circ}$ decreases by 38.2.

%


\section{Conclusions}
\label{sec:conclusion}

In this paper, we proposed \textit{TransNet}, a two-stage pipeline for category-level transparent object pose estimation. \textit{TransNet} outperformed a baseline by taking advantage of both state-of-the-art depth completion and opaque object category pose estimation. Ablation studies about multi-modal input and feature embedding modules were performed to guide deeper explorations. In the future, we plan to explore how category information can be used earlier in the network for better accuracy, improve depth completion potentially using additional consistency losses, and extend the model to be category-level across both transparent and opaque instances.

\clearpage
%
%
\bibliographystyle{splncs04}
\bibliography{egbib}
\end{document}